\documentclass{article}
\usepackage{spconf,amsmath,graphicx, setspace, fancyhdr}
\pdfoutput=1
\allowdisplaybreaks

\title{A\MakeLowercase{rchitectural configurations, atlas granularity and functional connectivity with diagnostic value in} A\MakeLowercase{utism }S\MakeLowercase{pectrum} D\MakeLowercase{isorder}}

\name{Cooper J. Mellema, Alex Treacher, Kevin P. Nguyen, Albert Montillo}
\address{University of Texas Southwestern Medical Center\\
Lyda Hill Dpt. of Bioinformatics, Dallas TX}

\renewcommand{\headrulewidth}{0pt}
\begin{document}

\maketitle

\begin{abstract}
\vspace*{-4mm}
\vspace*{-6mm}
Currently, the diagnosis of Autism Spectrum Disorder (ASD) is dependent upon a subjective, time-consuming evaluation of behavioral tests by an expert clinician. Non-invasive functional MRI (fMRI) characterizes brain connectivity and may be used to inform diagnoses and democratize medicine. However, successful construction of predictive models, such as deep learning models, from fMRI requires addressing key choices about the model's architecture, including the number of layers and number of neurons per layer. Meanwhile, deriving functional connectivity (FC) features from fMRI requires choosing an atlas with an appropriate level of granularity. Once an accurate diagnostic model has been built, it is vital to determine which features are predictive of ASD and if similar features are learned across atlas granularity levels. Identifying new important features extends our understanding of the biological underpinnings of ASD, while identifying features that corroborate past findings and extend across atlas levels instills model confidence. To identify aptly suited architectural configurations, probability distributions of the configurations of high versus low performing models are compared. To determine the effect of atlas granularity, connectivity features are derived from atlases with 3 levels of granularity and important features are ranked with permutation feature importance. Results show the highest performing models use between 2-4 hidden layers and 16-64 neurons per layer, granularity dependent. Connectivity features identified as important across all 3 atlas granularity levels include FC to the supplementary motor gyrus and language association cortex, regions whose abnormal development are associated with deficits in social and sensory processing common in ASD. Importantly, the cerebellum, often not included in functional analyses, is also identified as a region whose abnormal connectivity is highly predictive of ASD. Results of this study identify important regions to include in future studies of ASD, help assist in the selection of network architectures, and help identify appropriate levels of granularity to facilitate the development of accurate diagnostic models of ASD. 

\end{abstract}
\begin{keywords}
Interpretable deep learning, hyper parameter optimization, functional neuroimaging, fMRI, Autism.
\end{keywords}
\vspace*{-1mm}
\section{Introduction}
\label{sec:intro}

Autism Spectrum Disorder (ASD) is a common developmental disorder affecting 1 in 160 children annually and is characterized by abnormal neurological development\cite{DiMartino.2014}. Diagnosis of ASD currently consists of an extensive battery of behavioral tests which are evaluated by experts. These experts are not available at many clinics, hence accurate ASD diagnosis is less available than desired. Consequently there is growing interest in the development of an accurate, objective, fast, and reproducible diagnostic approach. One such approach uses functional MRI (fMRI) and structural MRI (sMRI) which can measure anatomical and functional alterations manifest in ASD\cite{Parisot.2018, Meenakshi2018}.  This approach is particularly promising when the imaging is used as the input to train a machine learning model to predict whether the subject has ASD or is a typically developing subject (e.g. a healthy control). Prior work has shown that such models can achieve between 70.4\% and 80.4\% area under the ROC curve (AUROC) \cite{Parisot.2018, Meenakshi2018, R.ToroN.TrautA.BeggatioK.HeuerandG.Varoquauxetal..2018, Mellema.2019}. 

This work
extracts information from the deep learning hyperparameter optimization, allowing us to identify configurations that lead to high performance and whether search ranges were adequate to isolate local performance maxima.
This work also determines the most important FC features used by each model through permutation feature importance and compares these to regions known to be affected by ASD, which would help grow confidence that the models have learned appropriately.
Finally, this work compares the discovered features across three levels of brain-atlas granularity. Features learned in common across models trained from different granularities can further corroborate their importance, and potentially identify novel features warranting further investigation.

\vspace*{-3mm}
\section{Methods}
\label{sec:format}

\vspace*{-3mm}
\subsection{Materials}
\vspace*{-3mm}

We trained models from 915 subjects from the IMPAC database\cite{R.ToroN.TrautA.BeggatioK.HeuerandG.Varoquauxetal..2018}. This included 418 subjects diagnosed as ASD and 497 identified as Healthy Controls (HC). The dataset includes one set of structural features (e.g. volumes and thickness of cortical regions) derived from structural MRI and several sets of functional features (inter regional connectivity) derived from resting-state fMRI (rs-fMRI). To derive the functional features, each subject was parcellated using one of 7 different atlases and the mean regional time signal was computed. Connectivity between these time courses was computed between pairs of regions using the tangent-space embedding metric\cite{Varoquaux.2011}.  

\vspace*{-3mm}
\subsection{Construction of DL models to predict ASD vs HC}
\vspace*{-3mm}
We conducted an extensive evaluation of 12 different machine learning (ML) models including 6 nonlinear ML classifiers, 3 linear ML classifiers, and 3 deep learning classifiers. To train the models, the subjects were initially randomly split 80\%/20\% into train and test partitions. The test data was held aside and not used until the final model evaluation. 

\vspace*{-3mm}
\subsection{Architecture optimization, model and atlas selection}
\begin{figure} [t!]
    \vspace*{-3mm}

  \includegraphics[width=8.5cm]{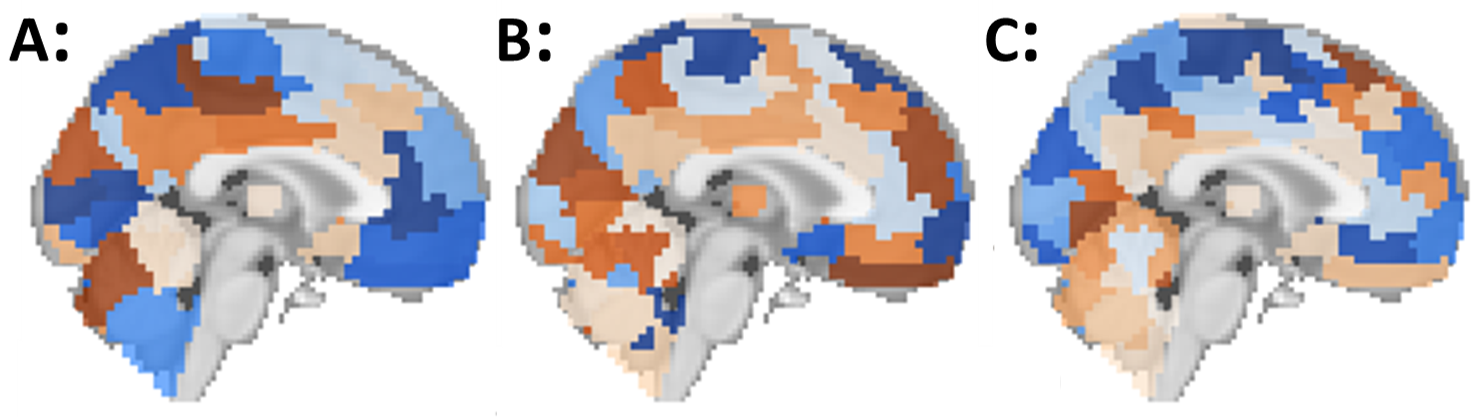}
    \vspace*{-3mm}

  \caption[example] 
  { \label{BASC Atlas} 
Brain parcellations (BASC \cite{Bellec.2010}) with varying granularity. (A) Coarse-grained with 64 ROIs, (B) medium-grained with 122 ROIs, (C) Fine-grained with 197 ROIs.}

\vspace*{-4mm}

  \end{figure}

\vspace*{-3mm}
To fairly evaluate the models and avoid biasing the results, 50 points in hyperparameter space were randomly chosen for each model and each of these 50 models were trained across 3 cross-validation folds from the 80\% training set.  The highest performing models were selected by mean AUROC across the cross-validation folds. For further details see \cite{Mellema.2019}. Our best models achieved 75.4-80.4\% ASD vs HC diagnosis accuracy using the BASC atlas \cite{Bellec.2010} as opposed to 4 other atlases tested at 3 levels of granularity. This atlas' coarsest resolution contains 64 ROIs (Fig. \ref{BASC Atlas}A), its medium-grained granularity has 122 ROIs (Fig. \ref{BASC Atlas}B), while its fine-grained granularity has 197 ROIs(\ref{BASC Atlas}C).  
Some of the highest, recently reported ASD vs. Control classification accuracies include those of Parisot\cite{Parisot.2018} who reports 70.4\% accuracy and Meenakshi\cite{Meenakshi2018} who reports 73.3\% accuracy.
In the present study, the \textit{DenseFFwd} models trained on each BASC atlas are the subject of analysis and interrogation because they performed well compared to these leading published results.

\vspace*{-3mm}
\subsection{High performing architectural configurations}
\vspace*{-3mm}

Hyperparameter searches generate a wealth of information. To gain insights from this information, kernel density estimates were computed for the models with the top 20\% of performance and for the models with the lowest 20\% of performance. This allows identification of architectural configurations that tend to produce high performing models and configurations that tend to produce low performing models. In addition, this analysis can indicate whether hyperparameter search ranges were adequate.

\vspace*{-3mm}
\subsection{Computation of feature importance}
\vspace*{-3mm}

The importance of each feature for each model was computing using permutation feature importance (PFI)\cite{Altmann}. In this approach, for a given trained model, each feature is permuted individually. Its feature importance, $I$,  is calculated as the z-score normalized mean decrease in AUROC: $I=P_b-P_a$, between the performance before feature permutation ($P_b$) minus the performance after feature permutation ($P_a$). PFI was chosen because it can be applied uniformly to different model and feature types. 
To aid in the comparison of IMPAC connectivity features to the literature which often reports results in Brodmann areas (BA), the centroid of each ROI for each atlas was calculated and matched to the corresponding BA for cross-study comparison. The ROI-ROI connection can then be re-written as the closest BA-BA connection.

\vspace*{-3mm}
\section{Results}
\label{sec:pagestyle}

  \begin{figure} [t!]
      \vspace*{-6mm}

  \begin{center}
  \begin{tabular}{c} 
  \hspace{-4mm}
  \includegraphics[width=8.5cm]{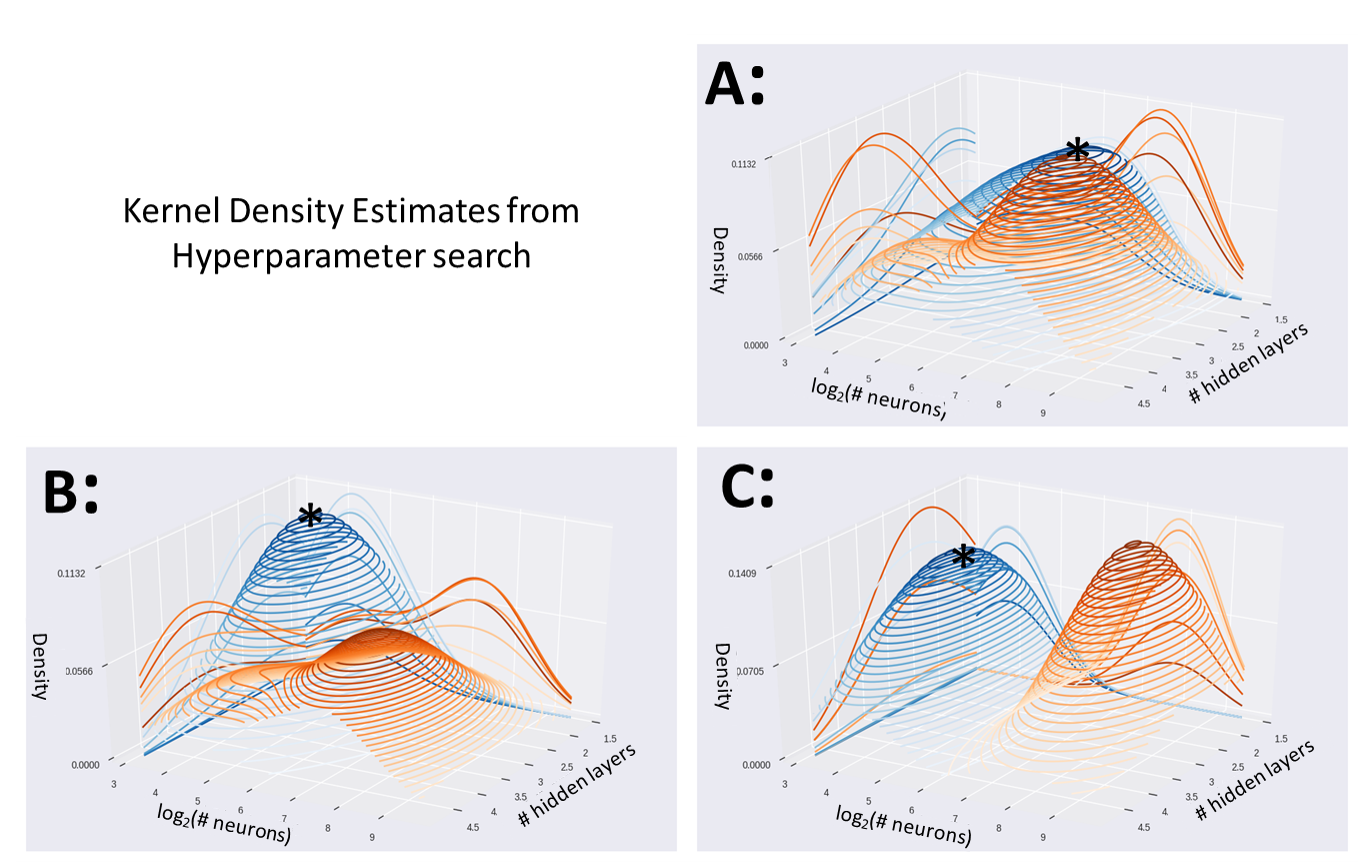}
        \end{tabular}
        \end{center}
    \vspace*{-7mm}

  \caption[example] 
  { \label{HyperParams} 
Kernel Density Estimates from the hyperparameter search reveals the density of top performing configurations (top 20\%) shown in blue, and low performing configurations (bottom 20\%) in orange. Densities of \textit{DenseFeedFwd} configurations using the coarse BASC atlas (A), medium atlas (B), and fine atlas (C). Peaks of blue surfaces are marked with *.}

    \vspace*{-2mm}

  \end{figure}

\thispagestyle{fancy}
\fancyhf{}
\renewcommand{\headrulewidth}{0pt}
\fancyfoot[CE,CO]{\textsf{\textbf{\fontsize{8}{12} \selectfont 1023}}}

\vspace*{-3mm}
Performance of a diagnostic predictive model can depend substantially on the choice of architectural configuration. In order to understand whether this is applies here, kernel density estimates were computed to estimate the probability distribution functions of the configurations of the top performing (top 20\%) configurations and bottom performing (bottom 20\%) configurations (Fig. \ref{HyperParams}).
Quantitatively, high performing \textit{DenseFFwd} models tended to use 1-2 hidden layers with 64 neurons per layer versus 3-4 layers with 256 neurons for the low performing models \textit{when using the coarse atlas}  (Fig. \ref{HyperParams}A), 2 layers with 16-32 neurons vs. 3 layers with 128 neurons for the \textit{medium-grained atlas} (Fig. \ref{HyperParams}B), and 3-4 layers with 16 neurons vs. 2 layers with 256 neurons for the \textit{fine-grained atlas} (Fig. \ref{HyperParams}C). As the peaks of the high (blue) and low (orange) performing models are not proximal, this suggests that \textit{configuration impacts performance substantially} (AUROC varied by ~20\% between high and low performing models). Also, we observe that the configurations of the top performing models, i.e. at the peaks in the blue surfaces, occur near the centers of the search ranges and not near the edges of the search space. This suggests that the search ranges have adequate coverage to discover good configurations. 

\begin{figure} [t!]
    \vspace*{-3mm}

  \includegraphics[width=8.5cm]{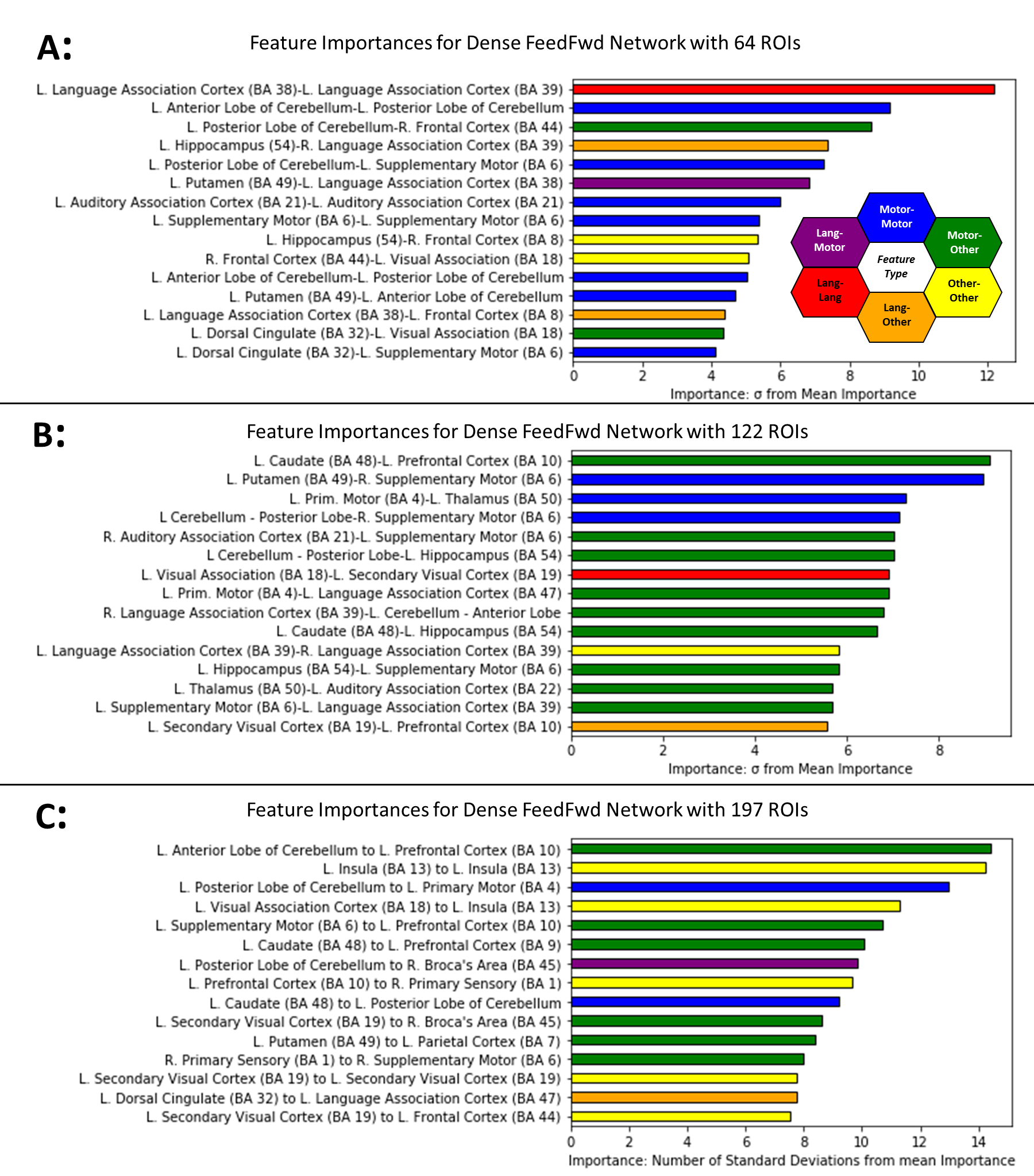}
    \vspace*{-2mm}

  \caption[example] 
  { \label{FunctionalFeatures} 
Important features for learned by highly accurate models for ASD diagnosis at each level of atlas granularity with coarse-grained atlas (A), medium-grained (B), and fine-grained (C). Each feature is the functional connectivity between two brain regions and is given a distinct color based on the function of these regions. Connections between sensorimotor ROIs are shown in blue, while connections between language ROIs are in red. Connections between regions that are neither motor nor language are in yellow. A connection  between language (red) and motor (blue) ROIs is shown with an intermediate hue (i.e. purple) and similarly for other region function combinations.
}
\vspace*{-3mm}

  \end{figure}

The top 15 features ranked by their feature importance for each atlas granularity are shown in Fig. \ref{FunctionalFeatures}. The feature importance for the connectivity features are reported as the number of standard deviations from the mean calculated feature importance (z-score). The most important features for the ASD vs HC prediction for the model trained with 64 ROIs is shown in Fig. \ref{FunctionalFeatures}A, whereas Fig. \ref{FunctionalFeatures}B and  Fig. \ref{FunctionalFeatures}C show  the most important features for the models trained from 122 and 197 ROIs respectively.  Color-coded functional labeling of features is shown to facilitate qualitative comparison. Motor, sensory, and language areas appear throughout the top features, while no structural features cCortical thickness, volume, etc.) were among the top 15 most important discriminative features.

The three most important features for the \textit{DenseFFwd} model trained from 64 ROIs were within the left language association cortex, within the left cerebellum, and between the left posterior cerebellum and right frontal cortex. For the model trained using the medium-grained atlas (122 ROIs), the top 3 features were between motor regions and the prefrontal cortex, and two within motor cortex alone. For the fine-grained atlas (197 ROIs), the top three features were between motor regions and the prefrontal cortex, within the insula, and between motor regions
. 

  \begin{figure} [t!]
      \vspace*{-3mm}

  \begin{center}
  \begin{tabular}{c} 
  \hspace{-3mm}
  \includegraphics[width=8.5cm]{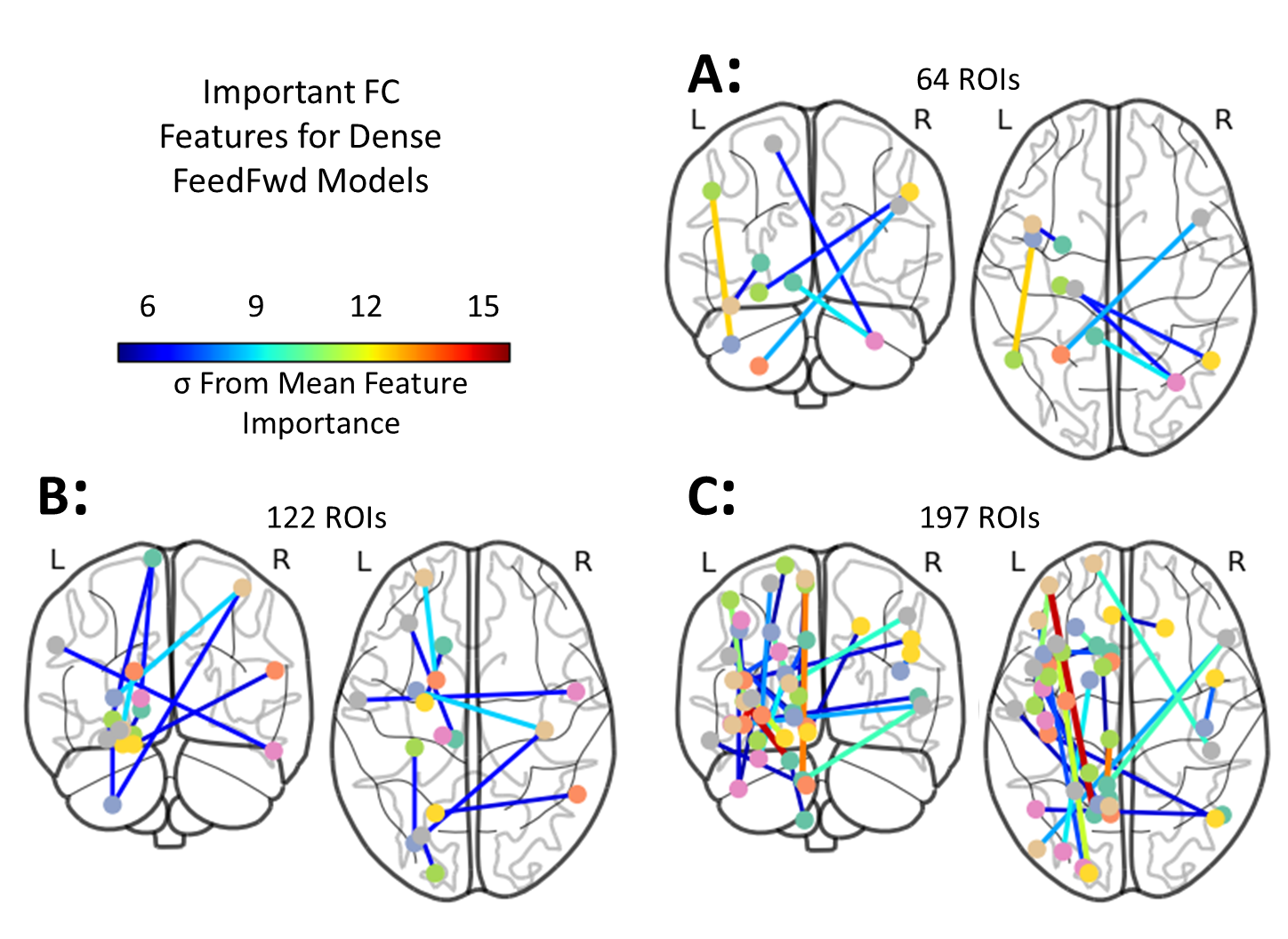}
        \end{tabular}
        \end{center}
    \vspace*{-7mm}

  \caption[example] 
  { \label{GlassBrainBiomarkers} 
Neuroanatomical locations of the most important functional connectivity features  and their relative importance. Features for the \textit{DenseFFwd} model using the BASC atlas with coarse (A), medium (B), and fine (C) granularities. Features with z score $\geq$6 are shown while their color indicates the number of standard deviations they are from the mean feature importance.}

\vspace*{-2mm}

  \end{figure}

The anatomical location of features with a z-score$\geq$6 are shown in Fig. \ref{GlassBrainBiomarkers}. There is substantial overlap in the features with high importance across all levels of granularity tested.

\vspace*{-2mm}
\section{Discussion}

\thispagestyle{fancy}
\fancyhf{}
\renewcommand{\headrulewidth}{0pt}
\fancyfoot[CE,CO]{\textsf{\textbf{\fontsize{8}{12} \selectfont 1024}}}

\label{sec:typestyle}
\vspace*{-3mm}

Our hyperparameter search analysis revealed that the highest performing models used between 2 and 4 hidden layers with 16-64 neurons per layer, with the optimal number of layers increasing with granularity and the optimal number of neurons/layer decreasing with increasing granularity.

Several regions were consistently found to have altered FC across the 3 accurate models we examined. When considering the top 15 most important features for each model, \textit{supplementary motor regions} (BA 6) were involved in 11 FC features across all three models, the \textit{posterior cerebellum} is involved in 8 FC features, and \textit{language association cortex (BA 39) and secondary visual cortex (BA 19)} are involved in 6 FC features across all three models. Similarly, the \textit{left anterior cerebellum}  is involved in 5 FC features, and Brodmann areas 54, 49, 48, 18, and 10 are all involved in 4 FC features across the three models. Additionally, we observe that many of the nodes implicated at a coarser resolution are also important at finer granularities (Fig. \ref{GlassBrainBiomarkers}). For example, the FC from the supplementary motor area to the cerebellum and itself are important in the coarse atlas, while the supplementary motor FC to the putamen, cerebellum, hippocampus, and prefrontal cortex, and sensory cortex are important in the fine grained atlas. Some of the differences in important edges across granularity may be explained by movement of the apparent regional centroid when a region is fine-grained. However, it also may be that these patterns of connectivity emerge only at specific scales. Further investigation into features variable across resolutions is warranted. That some features recurr at multiple resolutions bolsters confidence in their importance and suggests that even higher granularity may be warranted to further elucidate biological underpinnings. 

Many of the features identified here are in agreement with alterations reported previously including the significantly altered DMN connectivity, \cite{Jung.2014,Jung.2015,Uddin.2013}, connectivity in visual areas\cite{Chen.2015,Uddin.2013},  \cite{Chen.2015,DiMartino.2011,Rudie.2012}, motor and supplementary motor connectivity\cite{DiMartino.2011}, connectivity in somatosensory association areas\cite{Bhaumik.2018,Rudie.2012}, and connectivity in the prefrontal cortex\cite{Bhaumik.2018,Jung.2014,Jung.2015} in individuals with ASD. 
 
Importantly, our analysis also indicates that the FC with the cerebellum, including both the anterior and posterior aspects, are important diagnostic predictors. Moreover, these cerebellar features are important across \textit{all levels of granularity examined} (from the BASC atlas at 64, 122, and 197 ROIs). These consistent discriminatory connections lie between the cerebellum and motor areas as well as between the cerebellum and frontal cortex, regions that pertain to sensory processing and social behavior, well known to be altered in ASD. The altered FC between the cerebellum and frontal and sensorimotor cortices as a marker of ASD \textit{has received little attention in the literature}, as the cerebellum is often not included in functional analyses.  We suggest that these connections are areas worthy of further investigation and research.
 
There are several limitations in this study. First, this dataset only provides us a binary diagnosis and, as ASD is well-known to be a spectrum disorder, a dataset with finer gradiations of diagnosis would be expected to provide a more precise diagnostic model. Second, other methods than PFI should be applied to the models, such as layer-wise relevance propagation, to further explore the learned abstractions.

\vspace*{-3mm}
\section{Conclusion}

\thispagestyle{fancy}
\fancyhf{}
\renewcommand{\headrulewidth}{0pt}
\fancyfoot[CE,CO]{\textsf{\textbf{\fontsize{8}{12} \selectfont 1025}}}

\label{sec:typestyle}
\vspace*{-3mm}
In conclusion, this work has characterized the architectural configurations that lead to  high performing Deep Learning diagnostic models for ASD across 3 levels of granularity. This work has also identified the most important features using permutation feature importance analysis. The feature analysis identified new regions such as the anterior and posterior cerebellum with diagnostic importance and identified features in agreement with neuroanatomical regions previously implicated in ASD. That these regions were overlapping across 3 levels of regional granularity bolsters confidence that the models have discovered true discriminative features that may generalize well to new datasets and the clinic.  We look forward to extending this work with further development to include additional regions such as the brainstem, further testing, and clinical translation.

\vspace*{-2mm}


\bibliographystyle{IEEEbib}
\setstretch{0.05}
\bibliography{refs}
\end{thebibliography}

\end{document}